\documentclass[a4paper,fleqn]{cas-sc}

\usepackage[authoryear,longnamesfirst]{natbib}
\usepackage{pgfplots}
\usepackage{tabularx}
\usepackage{caption}
\usepackage{subcaption}
\usepackage{bbm}
\usepackage{url}
\usepackage{amsmath} 
\usepackage{pifont}
\usepackage{arydshln}
\usepackage{makecell}

\def\tsc#1{\csdef{#1}{\textsc{\lowercase{#1}}\xspace}}
\tsc{WGM}
\tsc{QE}
\tsc{EP}
\tsc{PMS}
\tsc{BEC}
\tsc{DE}
\begin{document}
\let\WriteBookmarks\relax
\def\floatpagepagefraction{1}
\def\textpagefraction{.001}

\shorttitle{Mitigating the Negative Impact of Over-association}

\shortauthors{Wang, Song, Min, Yao, Xu \& Su}

\title [mode = title]{Mitigating the Negative Impact of Over-association for Conversational Query Production}                      

\author[1,2]{Ante Wang}
\author[3]{Linfeng Song}
\author[1]{Zijun Min}
\author[4]{Ge Xu}
\author[1]{Xiaoli Wang}
\author[1]{Junfeng Yao}
\author[1,2]{Jinsong Su}
\cormark[1]

\affiliation[1]{organization={School of Information, Xiamen University},
    city={Xiamen},
    country={China}}
\affiliation[2]{organization={Xiamen Key Laboratory of Intelligent Storage and Computing},
    city={Xiamen},
    country={China}}
\affiliation[3]{organization={Tencent AI Lab},
    city={Bellevue},
    state={WA},
    country={USA}}
\affiliation[4]{organization={College of Computer and Control Engineering, Minjiang University},
    city={Fuzhou},
    country={China}}

\cortext[cor1]{Corresponding author}

\begin{abstract}
Conversational query generation aims at producing search queries from dialogue histories, which are then used to retrieve relevant knowledge from a search engine to help knowledge-based dialogue systems.
Trained to maximize the likelihood of gold queries, previous models suffer from the data hunger issue, and they tend to both drop important concepts from dialogue histories and generate irrelevant concepts at inference time.
We attribute these issues to the \emph{over-association} phenomenon where a large number of gold queries are indirectly related to the dialogue topics, because annotators may unconsciously perform reasoning with their background knowledge when generating these gold queries.%\footnote{This phenomenon is also highly related to the one-to-many phenomenon of dialogues, where there are usually various possible directions to continue a conversation.}
We carefully analyze the negative effects of this phenomenon on pretrained Seq2seq query producers and then propose effective instance-level weighting strategies for training to mitigate these issues from multiple perspectives.
Experiments on two benchmarks, Wizard-of-Internet and DuSinc, show that our strategies effectively alleviate the negative effects and lead to significant performance gains (2\%\,$\sim$\,5\% across automatic metrics and human evaluation).
Further analysis shows that our model selects better concepts from dialogue histories and is \emph{10 times} more data efficient than the baseline.
The code is available at \url{https://github.com/DeepLearnXMU/QG-OverAsso}.
%\footnote{Code will be released upon acceptance.}
\end{abstract}

% \includegraphics{figs/grabs.pdf}

% \begin{highlights}
% \item Research highlights item 1
% \item Research highlights item 2
% \item Research highlights item 3
% \end{highlights}

% Keywords

\maketitle

\section{Introduction}

Leveraging external knowledge has been proven to be important for dialogue response generation \citep{dinan2018wizard,zhou2018dataset,zhou2020kdconv,wang2021naturalconv}.
Along this line, exploring the Internet for external knowledge is gaining popularity because of its continually updated content and broad coverage on a variety of domains \citep{komeili2022internet,dusinc2022}.
To retrieve useful knowledge from the Internet, the task of query production is proposed to assemble search queries from dialogue contexts to effectively interact with search engines.
This task is crucial because the quality of generated queries directly affects the relatedness of retrieved knowledge to the current dialogue contexts and user intent.

\begin{figure}[t]
\centering
\includegraphics[width=0.5\textwidth]{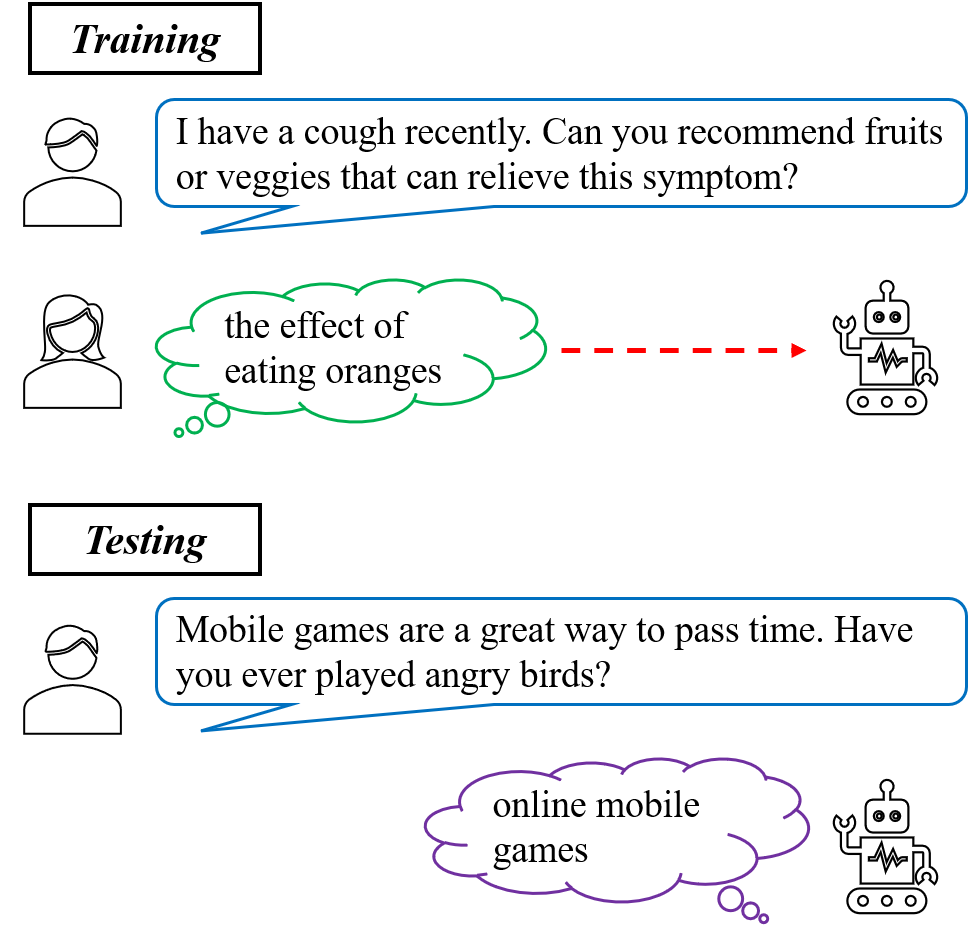}
\caption{Two typical conversations for query production. Trained with over-association instances such as the upper example, the query producer fails to predict the target topic ``\emph{angry birds}'' while generating irrelevant word ``\emph{online}''.}
% \caption{\revision{Two typical conversations for model training and testing, where utterances and queries are shown in boxes and bubbles respectively. The over-association query is highlighted with red bubbles. Training on a corpus suffering over-association, the model fails to predict the main topic about ``\emph{angry birds}'' when inference.}}
% \caption{One typical conversation with two alternative gold queries (\textcolor{darkgreen}{green} bubbles) from \emph{Wizard-of-Internet} \citep{komeili2022internet}, where the over-association part is highlighted in \textcolor{red}{\emph{italic}}. The corresponding model prediction (\textcolor{blueviolet}{purple} bubbles) is produced by a finetuned T5 model, which misses the important concept ``\emph{soccer}''.}
\label{fig:example}
% \vspace{-1.0em}
\end{figure}

The current efforts consider query production as a typical text-to-text generation task and adopt Transformer \citep{vaswani2017attention} as the backbone to build their query producers.
As a common practice, \citet{komeili2022internet} adopt a pretrained encoder-decoder model, such as BART \citep{lewis2020bart} or T5 \citep{raffel2020exploring}, and further finetune the model on a manually annotated query-generation dataset with the standard cross-entropy (CE) loss.
Benefited from a better initialization, the query producer can reach promising performance. % with limited training data.

However, we find that these query producers \citep{komeili2022internet,dusinc2022} still suffer from the data hunger issue as they usually require tens of thousands of training instances to reach decent performance.
Additionally, they tend to ignore important concepts from dialogue contexts and/or generate irrelevant concepts at inference time (``online mobile games'' in Figure \ref{fig:example}).
Our analyses show that these issues are related to the prevalent over-association phenomenon in the gold queries of major query generation benchmarks.
Particularly, a large number of gold queries contain new concepts absent from their dialogue contexts and not closely related to the current dialogue topics (``the effect of eating oranges
'' in Figure \ref{fig:example}).
We find that only 50.6\% gold queries in the \emph{Wizard-of-Internet} (WoI, \citealt{komeili2022internet}) development set can be assembled from dialogue contexts.
Trained with the standard CE loss to maximize the likelihood of each gold query, 
a query producer may show low confidence on the over-association parts of each query due to data sparsity,
causing large magnitude of loss values on these parts.
Consequently, the model requires a large number of training data to reach a stable performance, while it may still ignore crucial concepts from input dialogue contexts at inference time.
In summary, simply training a query producer as a standard text generation model may suffer from data hunger and unfaithfulness issues.

% Taking the case in Figure \ref{fig:example} for example, the dialogue context discusses playing soccer in the hot weather of San Antonio, while the annotated query asks general information on ``\emph{soccer positions}'', \revision{which start a indirectly related topic and involves a new concept ``\emph{positions}''.}
% which is not well related.

% \revision{
% Taking the case in Figure \ref{fig:example} for example, the \emph{user} asks \emph{bot} about its attitude to ``tom brady'', an American football quarterback. However, the annotated query for this turn is ``tom brady \emph{record}'', which is an excessive extension of the current topic and is not well related to the dialogue context.
% }

%The over-association phenomenon is inherited from the one-to-many property of open-domain dialogues, where there are usually various possible directions to continue a conversation.

In this paper, 
we first systematically analyze the over-association phenomenon and pinpoint its negative effects on several major pretrained models for text generation \citep{raffel2020exploring,lewis2020bart}.
We observe that training with over-association cases hurts model performance and decreases its overall confidence (even for simple cases) during inference.
Based on this observation, we then propose a novel model-agnostic training framework using \emph{data-based} and \emph{model-based} weighting strategies to mitigate the side effects of over-association cases.

Concretely, the data-based weighting strategy reduces the learning rate of a query according to its over-association degree defined based on how much of its content can be covered by the dialogue context.
% This strategy relies on the surface string matching between dialogue context and query to determine the over-association degree. 
However, it ignores model behaviors, which are also highly related to over-association based on our findings. Thus, we further propose the model-based weighting strategy. 
This strategy guides a model to learn from some of its own predictions (e.g., the step-wise model predictive probability or generated whole sequences).
% In particular, we explore two types of dynamic weighting by leveraging different model outputs:
% one only uses the 1-best model prediction, 
% while the other considers the model distribution of all possible predictions 
% (thus they are ``dynamic'' during training).
%Since the dynamic weighting may contain error, these two types unite gold query information to guide the model training.
Comparatively, the data-based weighting strategy mitigates the inconsistency between human annotations and dialogue contexts, while the model-based weighting strategy further mitigates the inconsistency between human annotations and model distribution.

Experiments on two benchmarks of different languages show that our model trained by either strategy generates more faithful queries than the baseline trained only with the CE loss.
And the combination of these two strategies leads to more performance gains.
% Our final model using either strategy shows additional gains in term of both human evaluation and automatic metrics, such as \emph{Unigram F$_{1}$} and \emph{BLEU-1/2}.
Further analyses in low-resource settings show that our model using just 10\% training instances reaches competitive performance with the baseline trained with the whole dataset.

\section{Research Objectives}
In this work, we aim to investigate the \emph{over-association} phenomenon of conversational query generation task. 
This phenomenon means that many manually labeled queries in training data are indirectly related to their dialogue contexts. 
Intuitively, a query producer trained on these instances can be misleading, thus it will drop important concepts or generate irrelevant concepts when inference.
Therefore, our research objectives are as follows:
\begin{itemize}
    \item Systematically analyze the negative impacts of the over-association phenomenon. We first define a metric to measure the over-association degree of each query. Then, we examine the behaviors of query producers trained on queries with different over-association degrees.
    \item Propose instance-wise weighting strategies to ease the negative impact of over-association. We explore improving the training objective term to solve this problem by decreasing the influence of search queries with high over-association degrees.
\end{itemize}

\section{Over-association in Conversational Query Production}
\label{sec:study}

In this section, we analyze the negative effects of the over-association phenomenon.
We take both T5 and BART as the baselines and evaluate them on two benchmarks in different languages, \emph{Wizard-of-Internet} (WoI, \citealt{komeili2022internet}) and \emph{DuSinc}\footnote{We use the part that is publicly available at \url{https://aistudio.baidu.com/aistudio/datasetdetail/139431/1}} \citep{dusinc2022}.
Following \citet{dusinc2022}, we report the Summation score (\emph{Sum.}) of \emph{Unigram F$_1$} and \emph{BLEU-1/2} as the evaluation metric in this section.\footnote{We use their released evaluation metrics described in \url{https://aistudio.baidu.com/aistudio/competition/detail/158/0/task-definition}. Detailed experimental settings can be found in \S \ref{sec:setup}}
We first introduce the baseline model (\S \ref{sec:baseline}), then present our observations (\S \ref{sec:analyse}). 

% \begin{table*}[t] \small
% \centering
% \begin{tabular}{lcccccccc|cccc}
% \toprule
% \multirow{3}{*}{Model} &\multicolumn{8}{c|}{Wizard-of-Internet} & \multicolumn{4}{c}{DuSinc} \\
% & \multicolumn{4}{c}{T5-base} & \multicolumn{4}{c|}{BART-base} & \multicolumn{4}{c|}{T5-base} \\
% & \ding{172} & \ding{173} & \ding{174} & ALL & \ding{172} & \ding{173} & \ding{174} & ALL & \ding{172} & \ding{173} & \ding{174} & ALL \\
% \midrule
% QP(\ding{172}) & 142.82 & 64.15 & 11.29 & 131.44 & 142.82 & 64.15 & 11.29 & 131.44 & 178.76 & 104.43 & 51.82 & 153.81 \\
% QP(\ding{172}+\ding{173}) & \textbf{144.49} & \textbf{65.63} & \textbf{11.51} & \textbf{133.38} & \textbf{144.49} & \textbf{65.63} & \textbf{11.51} & \textbf{133.38} & \textbf{179.62} & \textbf{107.39} & 68.14 & \textbf{156.42} \\
% QP(\ding{172}+\ding{174}) & 141.64 & 62.40 & 10.62 & 129.84 & 141.64 & 62.40 & 10.62 & 129.84 & 178.23 & 104.72 & 56.25 & 153.74 \\
% QP(ALL) & 142.48 & 65.28 & 11.35 & 131.65 & 142.48 & 65.28 & 11.35 & 131.65 & 177.22 & 103.54 & \textbf{69.10} & 154.17 \\
% \bottomrule
% \end{tabular}
% \caption{\emph{Sum.} on \emph{Wizard-of-Internet} and \emph{DuSinc} development set with different group combinations.}
% \label{tab:group}
% \end{table*}

\subsection{Baseline: Pretrained Text-to-text models with the Cross-Entropy Loss}
\label{sec:baseline}
Formally, taking a concatenated dialogue history $H=(h_1,...,h_n)$ of $n$ turns as inputs, the baseline query producer (\texttt{QP}) generates a search query $q=(q_1,...,q_m)$ of $m$ tokens.
Following previous work \citep{komeili2022internet}, the baseline model is initialized from either T5 \citep{raffel2020exploring} or BART \citep{lewis2020bart} before being finetuned on the query generation benchmark.
As an encoder-decoder model, the baseline model first consumes the concatenated dialogue history with its encoder, and then generates each query token $q_i$ given the previously generated tokens $q_{<i}$:
\begin{equation*}
p(q_i \mid H, q_{<i}; \theta) = \texttt{QP}(H, q_{<i}),
\end{equation*}
where $\theta$ represents the model parameters. This model is trained by maximizing the probability of the gold query with the standard CE loss:
\begin{equation}
\mathcal{L}_{CE} = - \sum_{i=1}^{m} \log p(q_i \mid H, q_{<i}; \theta).
\label{eq:loss}
\end{equation}

\subsection{Preliminary Study}
\label{sec:analyse}
To better understand the negative effects of over-association, we first propose a metric to estimate the over-association degree of a query considering its dialogue context.
Intuitively, a query suffering over-association usually can not be assembled from the dialogue history (e.g., ``\emph{the effect of eating oranges}'' in Table \ref{fig:example}). Therefore, we roughly measure the over-association degree of a query based on the rate of its word-level overlap with corresponding dialogue history:
\begin{equation}
    d(q) = 1 - \frac{\sum_{i}^{m} \mathbbm{1}_{H}(q_i)}{m} \text{,}
\label{eq:degree}
\end{equation}
where $\mathbbm{1}_{H}(q_i)$ is an indicator on whether the query word $q_i$ is contained by the dialogue history $H$.
Thus a higher score (lower word-level overlap) indicates that the query may suffer from a more serious over-association phenomenon.
To estimate it more reasonably, we remove the stop words and punctuations from the query and employ \emph{Spacy}\footnote{\url{https://spacy.io/}} to perform lemmatization for both dialogue history and query in advance. Notice that our defined automatic over-association degree is not completely accurate, but we find that it is highly related to the model performance (Pearson correlation coefficient 40.1\% for the over-association degree and \emph{Sum.} score of model prediction on the WoI development set).

Based on the over-association degree, we categorize the training instances into three datasets:

\begin{equation}
    \left\{\begin{array}{ll}
        q \in \text{\ding{172}} & d(q) \leq \frac{1}{3} \\
        q \in \text{\ding{173}} & \frac{1}{3} < d(q) \leq \frac{2}{3} \\
        q \in \text{\ding{174}} & \frac{2}{3} < d(q)
    \end{array}\right. \text{,}
\label{eq:dev_group}
\end{equation}
where datasets \ding{172}, \ding{173} and \ding{174} cover 74.6\%, 16.3\% and 9.1\% training instances of WoI, and the corresponding percents are 70.2\%, 21.8\% and 8.0\% for DuSinc.

\begin{figure}[t]
\centering
\includegraphics[width=0.5\textwidth]{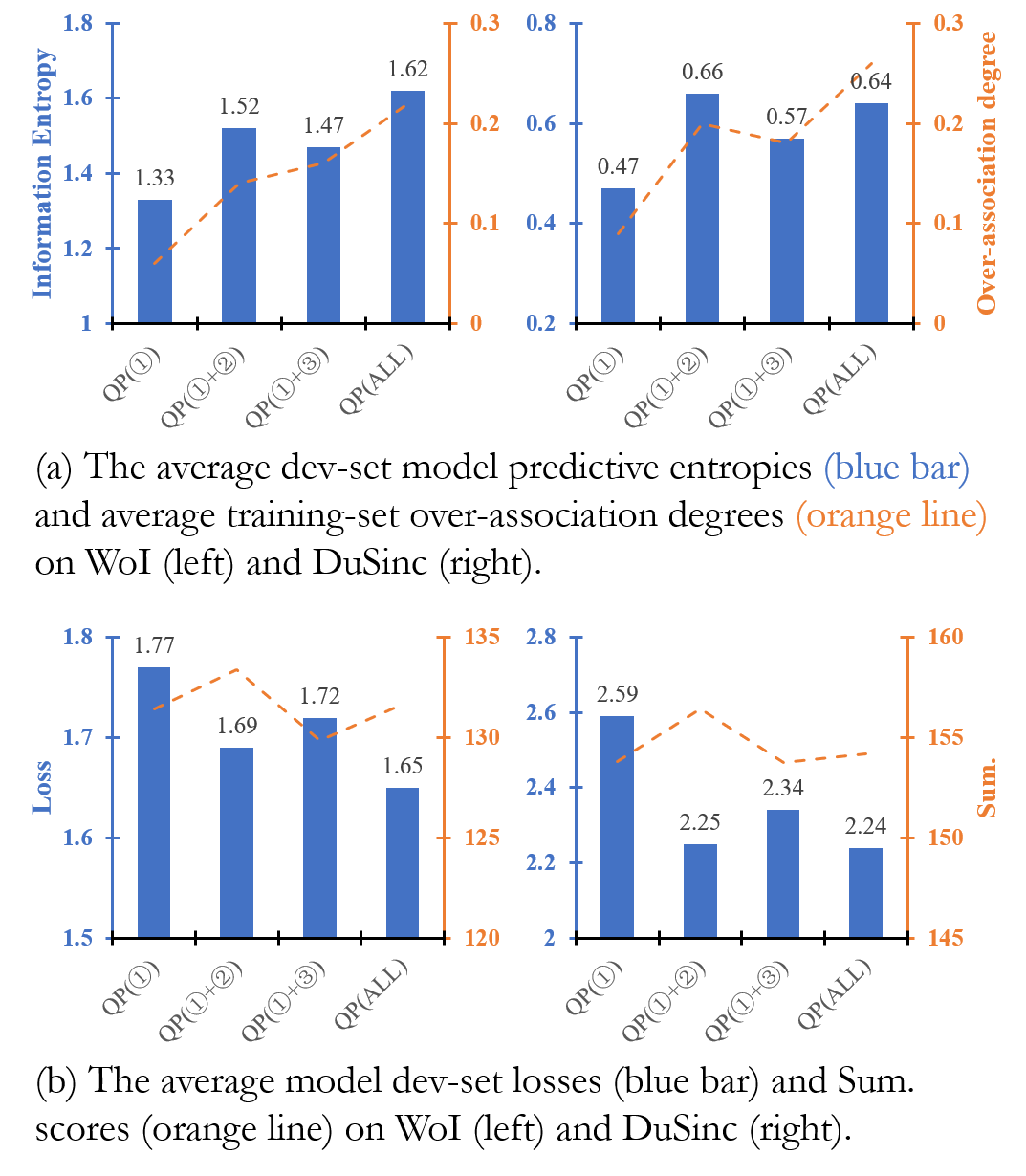}
% \vspace{-2.0em}
\caption{Analyses on the negative impacts of over-association.
}
\label{fig:entropy_loss}
% \vspace{-1.5em}
\end{figure}

Then, we conduct experiments that train the baseline models (\texttt{QP}) with the combinations of these datasets and analyze their evaluation metric scores (Table \ref{tab:group}), predictive information entropies (Figure \ref{fig:entropy_loss}(a)), and validation losses (Figure \ref{fig:entropy_loss}(b)) on the development sets. Particularly, four models are considered: 
\texttt{QP}(\ding{172}) is trained on the dataset \ding{172} and thus suffers from the least influences of over-association; 
\texttt{QP}(\ding{172}+\ding{173}) and \texttt{QP}(\ding{172}+\ding{174}) use more training instances with different degrees of over-association; \texttt{QP}(ALL) is trained on the whole dataset. Our conclusions are as follows.

\begin{table}[t] \scriptsize
\setlength\tabcolsep{5pt}
\centering
\caption{Summation score (\emph{Sum.}) of \emph{Unigram F$_1$} and \emph{BLEU-1/2} on different subsets (\ding{172}\,/\,\ding{173}\,/\,\ding{174}\,/\,ALL) from WoI and DuSinc development sets with different dataset combinations as training data.}
\begin{tabular}{lcc|c}
\toprule
\multirow{2}{*}{Model} &\multicolumn{2}{c|}{Wizard-of-Internet} &\multicolumn{1}{c}{DuSinc} \\ 
& T5-base (\ding{172}\,/\,\ding{173}\,/\,\ding{174}\,/\,ALL) & BART-base (\ding{172}\,/\,\ding{173}\,/\,\ding{174}\,/\,ALL) & T5-base (\ding{172}\,/\,\ding{173}\,/\,\ding{174}\,/\,ALL) \\
\midrule
\texttt{QP}(\ding{172}) & 142.82\,/\,64.15\,/\,11.29\,/\,131.44 & 130.26\,/\,62.28\,/\,13.07\,/\,112.73 & 178.76\,/\,104.43\,/\,51.82\,/\,153.81 \\
\texttt{QP}(\ding{172}+\ding{173}) & \textbf{144.49}\,/\,\textbf{65.63}\,/\,\textbf{11.51}\,/\,\textbf{133.38} 
& \textbf{130.28}\,/\,\textbf{65.21}\,/\,\textbf{17.06}\,/\,\textbf{113.14} & \textbf{179.62}\,/\,\textbf{107.39}\,/\,68.14\,/\,\textbf{156.42} \\
\texttt{QP}(\ding{172}+\ding{174}) & 141.64\,/\,62.40\,/\,10.62\,/\,129.84 & 128.10\,/\,63.16\,/\,16.39\,/\,111.13 & 178.23\,/\,104.72\,/\,56.25\,/\,153.74  \\
\texttt{QP}(ALL) & 142.48\,/\,65.28\,/\,11.35\,/\,131.65 & 126.56\,/\,63.84\,/\,16.34\,/\,109.99 & 177.22\,/\,103.54\,/\,\textbf{69.10}\,/\,154.17 \\
\bottomrule
\end{tabular}
\label{tab:group}
% \vspace{-1.0em}
\end{table}

As shown in Table \ref{tab:group}, 
more training data from \ding{173} or \ding{174} is not always leading to performance gains. 
For example,
\texttt{QP}(\ding{172}) performs worse than \texttt{QP}(\ding{172}+\ding{173}) in all settings. This indicates that more instances of moderate over-association degree can be beneficial.
However, \texttt{QP}(\ding{172}) and \texttt{QP}(\ding{172}+\ding{173}) outperforms \texttt{QP}(\ding{172}+\ding{174}) and \texttt{QP}(ALL), respectively, showing that \textbf{extreme cases in \ding{174} hurt the model performance} instead.

% \textbf{Moderate over-association degree data does benefit the model, but extreme ones hurt model performance.} As shown in Table \ref{tab:group}, \ding{172} performs worse than \ding{172}+\ding{173} but better than (or comparable with) \ding{172}+\ding{174}. Though introducing more training data, severe over-association data in \ding{174} still hurt a model's overall performance. And, it also explains why using all training data (i.e., \ding{172}+\ding{173}+\ding{174}) performs worse than \ding{172}+\ding{173}.

%% Second, comparing \texttt{QP}(\ding{172}) with \texttt{QP}(\ding{172}+\ding{174}) or \texttt{QP}(\ding{172}+\ding{173}) with \texttt{QP}(ALL) in Table \ref{tab:group}, directly fitting over-association cases mainly harms model performance on dataset \ding{172} or \ding{173}. In other words, \textbf{over-association cases mainly bring negative effects to the predictions of ones that are faithful to the dialogue context}.

% \textbf{Over-association cases mainly bring negative effects to the predictions of normal ones.} Comparing \ding{172} and \ding{172}+\ding{174} or \ding{172}+\ding{173} and \ding{172}+\ding{173}+\ding{174} in Table \ref{tab:group}, introducing training data in group \ding{174} mainly decrease model performance in group \ding{172} or group \ding{173} in which the cases are more faithful to dialogue contexts. It also indicates that we should not directly train the model with these data without distinction.

We choose T5-base for further study.
Figure \ref{fig:entropy_loss}(a) presents the entropy values of T5-base trained on data with different over-association levels. 
This is based on the assumption that if a model better fits the dataset, then the entropy of the model on it should be lower.
% Figure \ref{fig:entropy_loss}(a) presents the average entropies of model predictive distribution over vocabulary and average over-association degrees of training instances. 
In general, we observe a positive correlation between dev-set predictive entropy and training-set over-association degree, indicating that \textbf{training on the instances with higher over-association degrees often leads to lower model predictive confidence}.

% \textbf{Higher over-association degree leads to lower prediction confidence of a model.} Figure \ref{fig:entropy} presents the average over-association degree about training set and entropy of prediction probability. There is a clear negative correlation between them, which shows that over-association cases can make a model confusing.

% \begin{figure}[t]
% \centering
% \includegraphics[width=0.5\textwidth]{figures/loss.png}
% \caption{The average testing losses and \emph{Sum.} scores on WoI(left) and DuSinc(right) development sets for each model.
% }
% \label{fig:loss}
% \end{figure}

As shown in Figure \ref{fig:entropy_loss}(b), \texttt{QP}(ALL) reports significantly lower validation loss values than \texttt{QP}(\ding{172}), while this yields limited improvements on \emph{Sum.} score. Besides, \texttt{QP}(\ding{172}+\ding{173}) and \texttt{QP}(ALL) exhibit similar loss values but different \emph{Sum.} scores. These indicate that \textbf{there is a severe mismatch between the standard CE loss and the final model performance}.
Particularly, \texttt{QP}(ALL) shows much less validation loss values due to that it has used more instances during training, but it does not make fewer mistakes at inference time.
This further indicates the side-effect when we forcibly ask a model to perfectly fit the over-association instances.

%We also attribute this to the over-association phenomenon. 
%the model may produce a more flattened distribution to reach a lower CE loss, which instead hurts the model's predictive ability. It also meets our third conclusion above.

% \textbf{The drop of CE loss on the development set does not lead to performance gains.} As shown in Figure \ref{fig:loss},  \ding{172}+\ding{173}+\ding{174} significantly outperforms \ding{172} on loss, but have limited improvements on \emph{Sum.} score. \ding{172}+\ding{173} and \ding{172}+\ding{173}+\ding{174} have similar losses but much different \emph{Sum.} scores. This indicates that there is a severe mismatch between standard loss and the final performance of generated outputs.

\begin{figure}[t]
\centering
\includegraphics[width=0.5\textwidth]{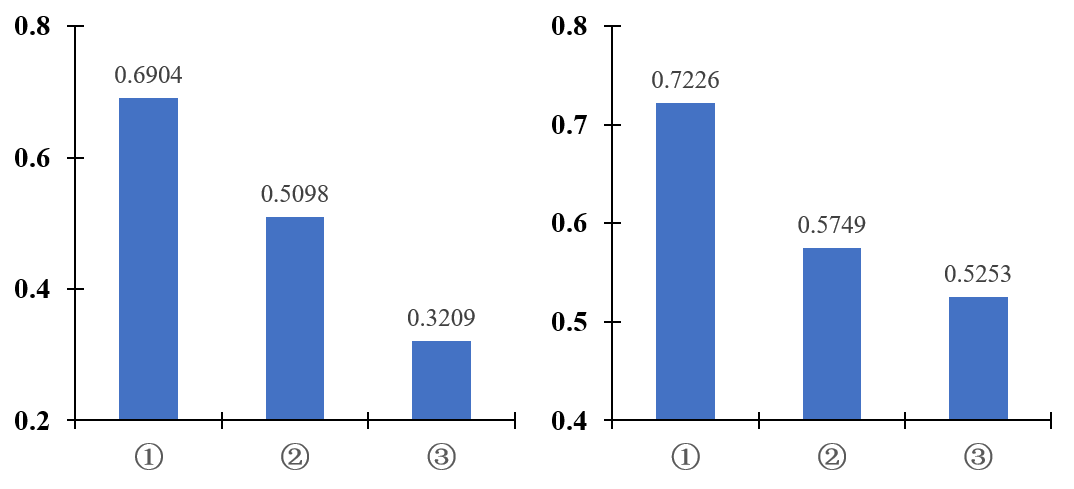}
\caption{Predictive probability of \texttt{QP}(ALL) to gold queries in different subsets of WoI (left) and DuSinc (right) development set.
}
\label{fig:prob}
\end{figure}

% \begin{figure}[t]
% \centering
% \includegraphics[width=0.6\textwidth]{figures/case_study_v2.png}
% \caption{An example where $Q_{1}$/$Q_{2}$ is a reference query with/without over-association, $P$ is a model generated query.
% }
% \label{fig:case_study}
% \end{figure}

Finally, we demonstrate that \textbf{the model behaviors (predictive probabilities and generated predictions) can somewhat reflect the over-association of input cases}.
In Figure \ref{fig:prob}, for \texttt{QP}(ALL), its predictive probability is always lower when testing on a dataset with more severe over-association. 
Besides, we have observed that the generated predictions from \texttt{QP}(ALL) always give higher matching scores (\textit{Sum.}) to non-over-association cases (Table \ref{tab:group}).
These findings inspire us to propose solutions to this problem in the later section.

\begin{figure*}[t]
\centering
\includegraphics[width=0.95\textwidth]{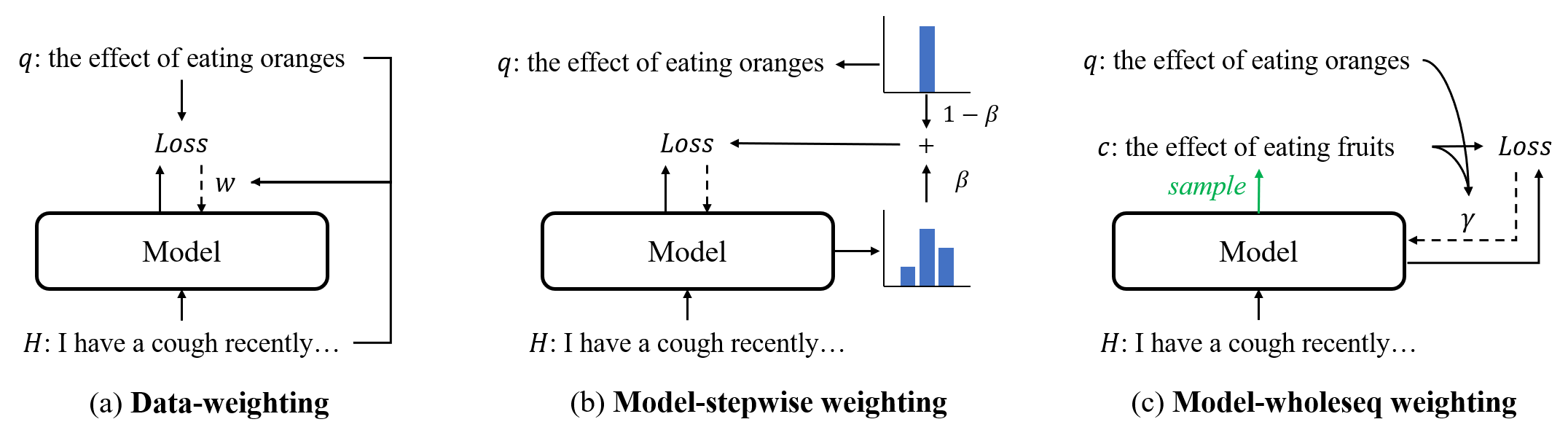}
% \vspace{-0.5em}
\caption{Three methods to tackle side effects of over-association.
}
\label{fig:model}
% \vspace{-0.5em}
\end{figure*}

\section{Mitigating the Negative Effects of Over-association}

Our analysis in \S \ref{sec:study} has shown the necessity to alleviate the side effects of over-association phenomenon. In this section, we propose two training strategies to alleviate the side effects: 1) \emph{data-based} weighting (\S \ref{sec:static}) guides the model with our automatic over-association degree. 2) \emph{model-based} weighting (\S \ref{sec:dynamic}) adjusts the training target considering some of the model outputs.
In addition to independent usage, the two strategies can also be combined, where data-based weighting is served as the warm-up step for model-based weighting.

\subsection{Data-based Weighting}
\label{sec:static}

As mentioned in \S \ref{sec:analyse}, we have explored \textbf{data-pruning}, which directly prunes the training instances with high over-association degrees, thus the side effects can be avoided.
% One simple method is \textbf{data-pruning}, which directly prunes the training instances with high over-association degree, thus the side effects can be avoided. It has been explored in our preliminary study (\S \ref{sec:analyse}). 
However, data-pruning is so strict that ignoring the harms from kept instances with moderate over-association degrees and the benefits from discarded ones. Thus we further propose \textbf{data-weighting} that carefully adjusts the learning rate of each instance.

As shown in Figure \ref{fig:model}(a), data-weighting applies a 0-1 scaling value to each training instance. The scaling value is decided by the over-association degree ($d(q)$ in Equation \ref{eq:degree}) for the instance. Formally, the scaling value $w$ is defined as
\begin{equation}
w = (1 - d(q))^{\alpha},
\end{equation}
where $\alpha$ is a hyperparameter determining the penalty to the over-association query. 
Finally, the loss function using the data-weighting becomes
\begin{equation}
\mathcal{L}_{weight} = - w \sum_{i=1}^{m} \log P(q_i \mid H, q_{<i}; \theta).
\label{eq:weight}
\end{equation}

Please note that a larger $\alpha$ denotes a smaller scaling value $w$ given a fixed over-association degree, meaning a larger penalty.
For instance, when over-association degree $d(q)=0.5$, increasing $\alpha$ from 1.0 to 2.0 yields a decrease of $w$ from 0.5 to 0.25.

\subsection{Model-based Weighting}
\label{sec:dynamic}

% The data-based weighting is mainly based on empirical findings: common tokens between the gold query and dialogue context are likely important.
% Nevertheless, there can still be a mismatch between the overlapping spans and the true important content.
% Besides, the over-association phenomenon may not be only confined to non-overlapping words. 
% A deeper reason can be that the current neural models are still incapable of fitting the human annotations (e.g. starting a new topic of ``soccer \emph{positions}'' in Figure \ref{fig:example}) suffering over-association.
% This large gap between human annotations and model distribution can lead to difficulty in achieving satisfying performance.

% In \S \ref{sec:analyse}, we have shown that the model trained on over-association cases tends to be unconfident (with higher entropy). 
% Besides, we further demonstrate that some kinds of model outputs (e.g., the predictive probability or generated predictions) are highly correlated with this problem.\footnote{Related analyses can be found in Appendix \S \ref{sec:evidence}.}

Though the data-based weighting strategy can ease the side effects to some extent, it can only capture the surface overlap between dialogue contexts and gold queries while ignoring the behaviors of the model.
%According to our analyses in Appendix \S \ref{sec:evidence}, we demonstrate that some model behaviors, like the predictive probability and generated predictions, are highly correlated with over-association.
Therefore, we further propose model-based weighting strategy, which considers either model stepwise predictive probability (\textbf{model-stepwise}) or generated whole sequences (\textbf{model-wholeseq}) to capture model behaviors as well.
This is inspired by our findings in \S \ref{sec:analyse} that both model behaviors can reflect over-association degrees of input cases from another aspect.
However, if asked to imitate these behaviors only, the model would suffer from its predicted noise.
Thus, both strategies we proposed jointly consider the model behaviors and gold queries, balancing the harms from the noise and the benefits from mitigating the side effects of over-association. 

\paragraph{Model-stepwise Weighting.} As shown in Figure \ref{fig:model}(b), this approach unites the original target vector and model predictive vector for each token in a query as the training target, which is also known as self-knowledge distillation \citep{hahn2019self,kim2021self,liu2021noisy}. The original target vector $\mathbf{y}_i$ is the $|V|$-dimensional one-hot vector of $q_i$, where $V$ is the vocabulary. The model predictive vector $P(H, q_{<i}; \theta)$ ($\mathbf{P}_i$ for clarifying) is the model output logit vector after the softmax normalization, which shares the same shape as $\mathbf{y}_i$. The model-stepwise weighting target for $q_i$ is the additive of $\mathbf{y}_i$ and $\mathbf{P}_i$ after scaling. Formally, the model is trained given by
\begin{equation}
\begin{aligned}
\mathcal{L}_{step} &= - \sum_{i=1}^{m} \left( \beta \mathbf{P}_i + (1 - \beta) \mathbf{y}_i \right)^{\mathrm{T}} \log \mathbf{P}_i, \\
% \mathcal{L}_{step} &= - \sum_{i=1}^{m} \sum_{y=1}^{|V|} P_{i,y} \log P(q_i=y \mid H, q_{<i}; \theta), \\
% P_i &= \beta \cdot P^e_i + (1 - \beta) \cdot P^t_i \text{,}
\end{aligned}
\label{eq:kd}
\end{equation}
% where $P_{i,y}$ is the target value of token $y$ in $P_i$. 
where $\beta$ is a hyperparameter and a large (small) $\beta$ denotes a favour to model (annotation) distribution.

Intuitively, $\mathcal{L}_{step}$ serves as a balance between trusting the model and entirely using the annotation when there is a large discrepancy between model predictive distribution and training target.
% One example can be predicting the next reference word ``\emph{oranges}'' given ``\emph{eating}'' in the upper example of Figure \ref{fig:example}, as the model may prefer to generate candidates like ``\emph{the effects of eating fruits}''.

%the target will loose the need for the model to generate over-association parts in annotations which are usually given low probabilities by the model. Meanwhile, it has less influence on the other non-over-association parts where the model is confident.

\paragraph{Model-wholeseq Weighting.} 
It is shown in Figure \ref{fig:model}(c) and updates model parameters in reinforcement learning style. 
The weight (i.e., reward) is used to ensure the quality of a generated prediction.
For each training instance, we first sample a candidate query $c$ from top-$\kappa$ generated predictions using beam search according to their predictive probabilities.
Then, we calculate the \emph{score of quality} $s$ for it against the gold query $Q$ using a scoring function $f$ (e.g., \emph{Unigram F$_1$}).
Both $\kappa$ and the scoring function $f$ are hyperparameters selected by development experiments.
Lastly, we adopt standard reinforcement learning \citep{williams1992simple} to update our model, and the loss is defined as
\begin{equation}
\mathcal{L}_{whole} = - \gamma \sum_{i=1}^{|c|} \log p(c_i \mid H, c_{<i}; \theta),
\label{eq:rl}
\end{equation}
where $c_i$ is the $i$-th token in $c$, 
and $\gamma$ is the reward for $c$ defined as $s - s_b$.
$s_b$ is the baseline reward designed to decrease the high variance of the gradient estimator, which is calculated by averaging the quality scores of top-$\kappa$ generated predictions.

Compared with model-stepwise weighting that only explores each annotated query, model-wholeseq weighting further expands to other candidates over the gold query.
Besides, model-wholeseq weighting deals with every whole candidate query, while model-stepwise weighting works on each decoding step.

\section{Experiment}

\subsection{Setup}
\label{sec:setup}

\paragraph{Dataset.}
We study on the following two benchmarks of query production in different languages.
% to better demonstrate this over-association phenomenon.
\begin{itemize}
\setlength\itemsep{0em}
    \item \textbf{Wizard-of-Internet} (WoI, \citealt{komeili2022internet}). It is a recently released English dataset that contains 35,765 / 2,534 / 2,164 pairs of dialogue contexts and queries in its training/development/test set. Each dialogue context corresponds to around 1.16 queries on average. For the test set, we merge instances according to the dialogue context and take the corresponding queries as multiple references.
    \item \textbf{DuSinc} \citep{dusinc2022}. It is a recent shared task on Chinese conversational query production containing 5,759 / 616 / 1,610 (dialogue context and query) pairs for training/development/test, respectively.
    %As it does not provide the test set during paper writing, we report the performance on the development set. %, taking the hyperparameters from WoI experiments.
\end{itemize}

\paragraph{Evaluation Metrics.}

We use the following automatic metrics to evaluate model performance:
\begin{itemize}
\setlength\itemsep{0em}
\item \textbf{Unigram F$_1$} (\emph{Uni. F$_1$}) measures the unigram overlap between predictions and gold references.
\item \textbf{BLEU-1/2} is a typical metric for text generation tasks. We use sacrebleu \citep{post2018call} for score calculation.
\item \textbf{ROUGE-1/2/L} \citep{lin2004rouge} is also a popular metric used for evaluating automatic summarization and machine translation.
\end{itemize}
We follow \citet{dusinc2022} to use \emph{Uni. F$_1$}, \emph{BLEU-1/2} and their summation (\textbf{Sum.}, the main metric for overall performance evaluation). The ROUGE scores will be provided in Appendix \ref{sec:rouge} as the supplementary evaluation.

\paragraph{Model Configuration.}
For the experiments on WoI, we use \emph{T5-v1.1-base} or \emph{BART-base}. For the experiments on DuSinc, we use \emph{mengzi-t5-base} \citep{zhang2021mengzi}, a strong model pretrained on a corpus derived from Chinese Wikipedia, Chinese News, and Common Crawl.\footnote{For all these pretrained models, we use the checkpoints from \url{https://huggingface.co/models}} 
All models are trained using Adam optimizer with the linear scheduler and initial learning rate of 5e-5. The batch size for models on WoI and DuSinc are 64 and 16, respectively. We train all models until convergence and select the checkpoints with best \emph{Sum.} score on development set as final models.

% \subsection{Setup}
% The basic training settings of static weighting and dynamic weighting follows our baseline in \S \ref{sec:baseline}. 
Specially, for the model-based weighting, we first pretrain the model using the standard CE loss or data-based weighting strategy. It ensures the quality of model predictive probability or generated predictions at the beginning.
The selection of hyperparameters $\alpha$, $\beta$, and $\kappa$ for WoI/DuSinc are 2.0/0.5, 1.0/0.75, and 10/10 respectively given development experiment results. 
We directly take \texttt{QP}(\ding{172}+\ding{173}) for data-pruning as it performs best in our preliminary study (\S \ref{sec:analyse}).
For scoring function $f$, we consider \emph{Uni. F$_1$}, \emph{BLEU-2} and semantic similarity score calculated by Sentence-BERT \citep{reimers2019sentence}. We select \emph{Uni. F$_1$} as $f$ considering performance and cost. Related experiment results are shown in Appendix \S \ref{sec:dev}.

\begin{table*}[t] \small
\setlength\tabcolsep{6pt}
\centering
\caption{Our main test results on the two datasets. \dag\;denotes the results from a previously released model. Data-based, model-based weighting strategies and their combination are separated with dashed lines.}
\begin{tabular}{lcccc|cccc}
\toprule
\multirow{2}{*}{Model} &\multicolumn{4}{c|}{Wizard-of-Internet} & \multicolumn{4}{c}{DuSinc} \\
& Uni. F$_1$ & BLEU-1 & BLEU-2 & Sum. & Uni. F$_1$ & BLEU-1 & BLEU-2 & Sum. \\
\midrule
Blenderbot2\dag & 43.27 & 37.59 & 30.76 & 111.62 & N/A & N/A & N/A & N/A \\
T5-base & 44.57 & 41.88 & 33.26 & 119.71 & 60.67 & 51.27 & 47.25 & 159.19 \\
T5-large & 46.80 & 44.56 & 35.26 & 126.62 & 64.36 & 54.15 & 50.10 & 168.61 \\
\midrule
Data-pruning & 45.09 & 42.28 & 33.59 & 120.96 & 61.30 & 51.35 & 47.26 & 159.91 \\
Data-weighting & 46.55 & 43.25 & 35.14 & 124.94 & 61.54 & 52.42 & 48.43 & 162.39 \\
\hdashline
Model-stepwise & 45.43 & 43.57 & 34.54 & 123.54 & 60.63 & 52.23 & 47.93 & 160.79 \\
Model-wholeseq & 48.19 & \textbf{46.22} & 36.45 & 130.86 & 63.71 & 53.13 & 49.06 & 165.90 \\
\hdashline
% \makecell[l]{Data-weighting$\rightarrow$\\Model-stepwise} & 46.36 & 43.93 & 35.40 & 125.69 & 61.90 & 53.40 & 49.25 & 164.55 \\
\makecell[l]{Data-weighting$\rightarrow$\\Model-wholeseq} & \textbf{48.90} & 46.04 & \textbf{36.69} & \textbf{131.63} & \textbf{63.86} & \textbf{53.75} & \textbf{49.80} & \textbf{167.41} \\
\bottomrule
\end{tabular}
\label{tab:main_exp}
% \vspace{-0.5em}
\end{table*}

\subsection{Main Results}

Table \ref{tab:main_exp} shows the overall results on WoI and DuSinc,
where \emph{BlenderBot2} represents the query generation model\footnote{We use their released checkpoint from \url{https://parl.ai/projects/blenderbot2}} by \citealt{komeili2022internet}.
\emph{BlenderBot2} is based on BART-large \citep{lewis2020bart} and is trained on the same data split as our models.\footnote{Actually we use their data split as mentioned in \S \ref{sec:setup}.}
All our models are based on T5-base, thus they all take smaller amounts of parameters than \emph{BlenderBot2} (220M vs. 406M).
% We also include \emph{Data-pruning} (i.e., \texttt{\ding{172}+\ding{173}}) proposed in \S \ref{sec:study} as a special approach of data-based weighting, which has shown best performance among different training group combinations.

We can draw the following conclusions.
\textbf{First}, 
we are surprised to find that \emph{T5-base} performs better than \emph{BlenderBot2}, even though \emph{T5-base} takes a smaller amount of parameters.
Except for the difference between their pretraining styles, one possible reason may be the metric used for checkpoint selection: we use \emph{Sum.} score instead of perplexity adopted by \emph{BlenderBot2}, given our finding that the CE loss (highly related to perplexity) and \emph{Sum.} are not well-related in \S \ref{sec:analyse}.
Nevertheless, this also indicates that our comparisons are based on a strong baseline.
\textbf{Second}, all models of either the data-based weighting or the model-based weighting outperform the baselines on both datasets, indicating the effectiveness and generalization of all methods for mitigating the negative effects of the over-association.
% For the models using static weighting, \emph{Weighting} scores higher on \emph{Wizard-of-Internet}, while \emph{Data-pruning} performs better on \emph{DuSinc}. It may be because \emph{Wizard-of-Internet} is 5.7 times larger than \emph{DuSinc} and a \emph{Weighting} model with limited training data is still not robust to resist over-association negative effects. Combining both \emph{Data-pruning} and \emph{Weighting} may tackle this problem, but it needs much more fussy hyper parameters selection.
For the models using data-based weighting, 
\emph{Data-weighting} performs better than \emph{Data-pruning} because it provides more sophisticated control for learning the over-association cases.
For the models using model-based weighting, 
% comparing with static weighting, \emph{Annotation-oriented} performs better on \emph{DuSinc}, but slightly worse on \emph{Wizard-of-Internet}. This is as expected because they function similarly by weighting the training instances but different granularity. However, \emph{Annotation-oriented} gets rid of handcraft feature (i.e., over-association degree) thus is easier to be applied in various settings.
\emph{Model-stepwise} is superior to \emph{Data-pruning} but inferior to \emph{Data-weighting}. It may be because the model-stepwise weighting provides control at a finer granularity (i.e., token-level), which is not suitable for a generation task taking a sequence of tokens as the reference.
\emph{Model-wholeseq} significantly outperforms other approaches. By not fitting the training instances directly, it benefits from the most \emph{flexible} supervision signal and \emph{minimal} constraint on fitting over-association cases. 
\textbf{Finally}, 
% we also explore combining both advantages of data-based and model-based strategies, where a model trained in data-weighting strategy is used as the model initialization before training in model-wholeseq strategy. 
we report \emph{Data-weighting$\rightarrow$Model-wholeseq}, which simply combines the stronger weighting strategies in data-based and model-based strategies, respectively. 
It gives the best results among all models and has more gains on DuSinc than WoI. 
This may be because \emph{T5-base} can provide promising initialization parameters with a fair amount of training instances (WoI is 5.7 times of DuSinc). We will further discuss it in low-resource experiments.
As an unfair setup, we also compare our models with baselines using T5-large\footnote{We use \emph{T5-v1.1-large} and \emph{mt5-large} for WoI and DuSinc, respectively.}, which are \emph{3.5} times larger than our models. 
\emph{Data-weighting$\rightarrow$Model-wholeseq} show slightly inferior results on DuSinc (-1.2 on \emph{Sum.}) but still give much better results on WoI (+5.01 on \emph{Sum.}).
This again strongly validates the effectiveness of our approach.

\subsection{Analysis}

In this section, we conduct more detailed comparisons between our strategies and the \emph{T5-base} on WoI to validate how we address the negative effects of over-association.
We also explore model performances in low-resource settings, as this is important in real-world scenarios.
We mainly report \emph{Data-weighting$\rightarrow$Model-wholeseq} (which is named as \emph{Combine} for simplicity), but also \emph{Data-weighting} and \emph{Model-wholeseq} as they perform stronger within data-based and model-based weighting strategies respectively.
% Note that we select \emph{Data-weighting} and \emph{Model-wholeseq} weighting from data-based and model-based weighting strategies in the remaining experiments because they report the strongest performances in the main test results respectively.

\begin{figure}
\centering
\begin{tikzpicture}[scale=0.8]
\hspace{-0.3cm}
\begin{axis}[
	xlabel=Training instance ratio,
	ylabel=Sum.,
	xtick={0,2,5,12},
	xticklabels={5\%, 10\%, 20\%, 100\%},
    legend style={
    nodes={scale=0.75, transform shape},
        at={(0.98,0.3)},
        cells={anchor=west}}
]
\addplot[dash dot, color=blue, mark=square*] coordinates {
        (0, 107.95)(2, 112.97)(5, 117.24)(12,119.71)
        };
\addlegendentry{T5-base}
\addplot[dashed, color=green, mark=triangle*] coordinates {
    (0, 114.46)(2, 118.19)(5, 119.79)(12,124.94)
    };
\addlegendentry{Data-weighting}
% \addplot[dotted, color=purple, mark=diamond*] coordinates {
%     (0, 110.29)(2, 115.73)(5, 119.33)(12,123.54)
%     };
% \addlegendentry{Model-stepwise}
\addplot[dashdotdotted, color=orange, mark=star] coordinates {
    (0, 114.12)(2, 116.93)(5, 125.23)(12,130.86)
    };
\addlegendentry{Model-wholeseq}
% \addplot[dashdotdotted, color=orange, mark=otimes] coordinates {
%     (0, 115.30)(2, 119.92)(5, 121.41)(12,125.69)
%     };
% \addlegendentry{\makecell[l]{Combination(Prob)}}
\addplot[sharp plot, color=red, mark=*] coordinates {
    (0, 114.98)(2, 119.17)(5, 125.43)(12,131.63)
    };
\addlegendentry{Combine}
% \addplot[loosely dashed, color=gray] coordinates {
%         (0, 119.71)(2, 119.71)(5, 119.71)(12,119.71)
%         };
\end{axis}
\end{tikzpicture}
% \vspace{-0.5em}
\caption{Model performances on WoI test set with different number of instances for training.}
\label{fig:data_hunger}
% \vspace{-0.5em}
\end{figure}
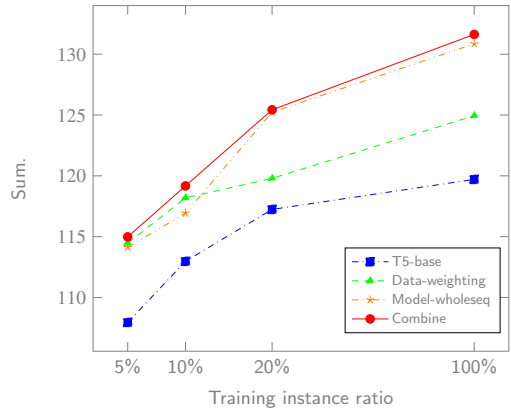

\paragraph{Mitigating the side effects of over-association eases data hunger issue.}
Because of the over-association, a model can need massive data to well fit the ground-truth answers.
This is a typical data hunger issue.
Figure \ref{fig:data_hunger} compares the \emph{Sum.} scores of our models and the baseline when trained with different ratios (5\%, 10\%, 20\%, 100\%) of the whole training corpus.
When training instances are limited, a model may be not robust enough to resist the negative effects. By mitigating these side effects, our methods require less data to train a model with decent performance.
Particularly, we observe the significant performance gain when using limited training data adopting either the data-based or model-based weighting strategy.
More importantly, 
using 10\% training instances, \emph{Combine} has performed comparably with the baseline trained with the whole dataset. And when using 20\% instances, it performs significantly better.
% using 20\% training instances, \emph{Data-weighting} performs comparably with the baseline trained with the whole dataset and \emph{Model-wholeseq} performs significantly better than it.
Interestingly, \emph{Model-wholeseq} is slightly worse than \emph{Data-weighting} when using 5\% or 10\% data. It is due to the unsatisfying initialization of model parameters, which meets the common practice: pretraining is important for reinforcement-learning-based finetuning \citep{li2016deep,paulus2018a}. 
% Thus, with a better initialization, the combination of both types of strategies gives better performances especially when the number of training instances is limited.
% Nevertheless, this experiment demonstrates the necessity of tackling the side effects of over-association and the efficiency of our approach to ease the data hunger issue.

\begin{figure}[t]
\centering
\includegraphics[width=0.5\textwidth]{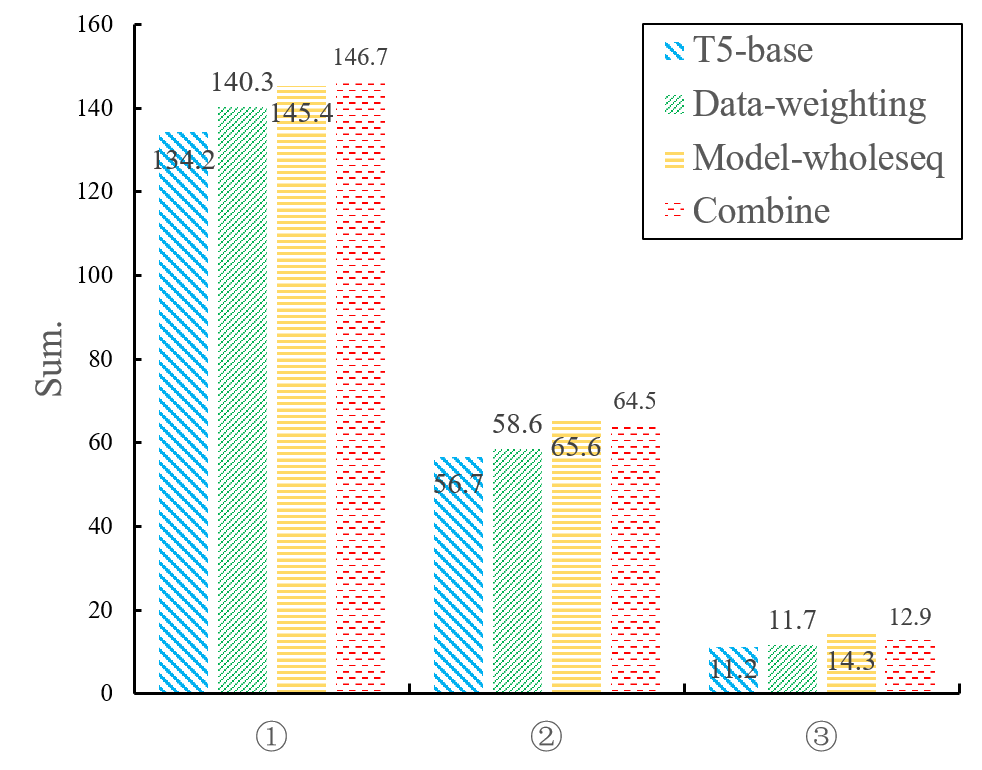}
% \vspace{-0.5em}
\caption{Performances over different subsets (\ding{172}, \ding{173} and \ding{174}) from WoI test set.
}
\label{fig:group}
% \vspace{-0.5em}
\end{figure}

\paragraph{Improvements for different over-association degrees.}
Similar to \S \ref{sec:analyse}, we split the test set into 3 subsets according to the over-association degree defined in Equation \ref{eq:degree}: \ding{172}, \ding{173} and \ding{174}. They cover 76.1\%, 15.9\%, and 8.0\% of the original test set, respectively.
Figure \ref{fig:group} shows the comparison of \emph{T5-base} and our models for each subset.
Generally, the performance of all models shows a clear downward trend with over-association degree increasing. 
Compared with \emph{T5-base}, both data-weighting and model-wholeseq weighting benefit all subsets even for \ding{174} with the most severe over-association. 
% This again confirms our motivation.
Besides, we notice that with a smaller over-association degree, the improvement of our models over \emph{T5-base} model is also more significant. 
When combining the two strategies, \emph{Combine} gives further improvements on \ding{172} but is slightly worse on \ding{173} and \ding{174}.
It shows that our methods mainly help a model perform better on faithful cases than the over-association ones.
This meets our expectations as our approach aims to alleviate the side effect of over-association training instances.

\subsection{Human Evaluation}

\begin{table}[t] \small
\centering
\setlength\tabcolsep{15pt}
\caption{Human evaluation (score scale from 0 to 2) on WoI test set.}
\begin{tabular}{lc}
\toprule
Model & Human Score \\
\midrule
T5-base & 1.60 \\
Data-weighting & 1.63 \\
Model-wholeseq & 1.71 \\
Combine & \textbf{1.73} \\
\bottomrule
\end{tabular}
\label{tab:human}
% \vspace{-0.5em}
\end{table}

Because of the high variance of annotations, automatic evaluation may not reasonably measure the performance of different models. Thus, we conduct a quantitative human study. 
Specifically, we sample 200 dialogue context and query pairs from WoI test set to conduct this study, asking 3 annotators capable of fluent English communication to score with 3-point schema (e.g. 0, 1 and 2) based on the soundness of each model-generated query.\footnote{Detailed guidelines are listed in Appendix \S \ref{sec:guideline}}
The inner-annotator agreement (Fleiss' $\kappa$) is 0.75, which is at the Substantial level.
We average the 3 scores for each instance, and scores of all instances are averaged as the final score for the corresponding model. 
Table \ref{tab:human} shows the results where our final model scores 0.13 points higher than the baseline.
This is roughly a 6.5\% gain, confirming the effectiveness of our method.

\subsection{Case Study}
\begin{table}[t!] \small
    \centering
    \caption{Two examples with their predictions of different models from WoI test set.}
    \begin{tabularx}{0.85\textwidth}{rX}
        \textbf{\#1} \\
        \midrule
        Dialogue & \emph{User}: I understand that grip strength is considered to be an indicator of longevity, and I'd like to write an article for my local paper about it, but I need some citations. Can you help me? \\
        & \emph{Bot}: Sure! The lancet had a study that indicated grip strength correlates with heart health. They think it may be better than reading your blood pressure at measuring your health. \\
        & \emph{User}: Wow! That's terrific. What year was that study? \\
        \midrule
        Gold Query & is grip strength an indicator of longevity \\
        \hdashline
        {\scriptsize T5-base} & lancet study \\
        {\scriptsize Data-weighting} & lancet grip strength \\
        {\scriptsize Model-wholeseq} & lancet grip strength study \\
        {\scriptsize Combine} & lancet grip strength study \\
        \midrule
        \textbf{\#2} \\
        \midrule
        Dialogue & \emph{Bot}: Are you more into easy or complex brews? \\
        & \emph{User}: Right now I'm just starting with easy brews. I'm not even sure what a complex brew is! \\
        \midrule
        Gold Query & complex coffee brew \\
        \hdashline
        {\scriptsize T5-base} & easy brews \\
        {\scriptsize Data-weighting} & easy brews \\
        {\scriptsize Model-wholeseq} & complex brew \\
        {\scriptsize Combine} & complex brews \\
        \bottomrule
    \end{tabularx}
    \label{tab:case_study}
% \vspace{-0.5em}
\end{table}

As shown in Table \ref{tab:case_study}, we further demonstrate two typical examples from our human study to help visualize the benefits of our approach on alleviating the negative effects of over-association phenomenon.
% The gold queries of both examples contain high variations. 
Both cases suffer from over-association.
For the first example, the \emph{User} asks ``\emph{What year was the lancet study}'', which indicates ``\emph{grip strength correlates with heart health}''.
Compared with the gold query (``\emph{is grip strength an indicator of longevity}''), we believe the query by \emph{Combine} (``\emph{lancet grip strength study}'') is more related to the context.
The baseline generates ``\emph{lancet study}'', which does not give enough details, as it can refer to other studies by \emph{lancet}. \emph{Data-weighting} also gives a more concrete result, while it misses the keyword ``study''.

For the second example, generating the word ``coffee'' needs guessing with commonsense knowledge. All models fail to predict ``coffee''. However, our \emph{Model-wholeseq} and \emph{Combine} are more \emph{accurate} that both of them correctly predict the following topic about ``complex brew''.

\section{Related Work}
%One side effect of Over-association is the unfaithfulness issue, where the model predictions fail to be consistent with model inputs.
%One direction to alleviate it is taking an additive reconstruction loss of inputs, which has been widely studied on sequence labeling tasks, such as named entity recognition \citep{rei2017semi,liu2018empower,jia2019cross}, machine translation \citep{tu2017neural} and data-to-text generation \citep{wiseman2017challenges,song2020structural}. 
%Different from these approaches, we attribute this issue in conversational query production to over-association. Thus we design several strategies to mitigate this phenomenon, simultaneously addressing the unfaithfulness issue. 
% We try to alleviate a similar unfaithfulness issue, but we focus on constructing the targets (instead of the inputs) in different methods to mitigate the side effects of over-association.
% Besides, we are the first to investigate such an idea in conversational query production.

Conversational query production is a fundamental task for building search-engine-augmented dialogue systems.
To accelerate the study of this research line, \citet{komeili2022internet} and \citet{dusinc2022} have contributed Wizard-of-Internet and DuSinc respectively, and trained their models in standard supervised learning.
As labeled queries are costly to obtain, \citet{WangSLMWTSY23} proposed to train their query producer in a weak-supervised manner by leveraging feedback from a search engine.
More recently, \citet{WangSXS23} and \citet{huang2023response} further enhanced their models using semi-supervised learning to tackle low-resource and domain adaptation challenges.
Differently, our method does not rely on any external resources thus we believe our method is also orthogonal to these concurrent approaches.

Our model-stepwise weighting is inspired by Self-knowledge distillation (SKD, \citealt{hahn2019self,kim2021self,liu2021noisy}), which is adopted as a regularization term to improve model generalization in both Computer Vision and Natural Language Processing fields.
Typically, previous work adopt SKD to tackle the over-confidently predicting problem or resist outliers in a noisy corpus.
Different from these efforts, we are the first to investigate SKD on query production
to alleviate the challenge caused by the over-association phenomenon.
%, which is highly related to the one-to-many property in dialogue modeling.
%While in this work, we aim to use model predictive probability to distinguish over-association terms and mitigate their negative impacts.

Reinforcement learning is a popular technique to solve the inconsistency between token-level training objective functions and sequence-level evaluation metrics on Seq2Seq generation tasks \citep{ranzato2015sequence,bahdanau2016actor,wu2018study}.
In this work, our model-wholeseq weighting mainly focuses on leveraging model predictions to exclude over-association cases instead of easing the inconsistency between training and testing. As model predictions may contain errors, we design the reward term using annotations to ensure the quality of training targets.

Conversational query production is remotely related to keyphrase generation \citep{meng2017deep,chen2018keyphrase,Chen_Gao_Zhang_King_Lyu_2019,chan2019neural,yuan2020one}, which aims at producing keyphrases that summarize the main topics of research articles.
% As most articles have multiple keyphrases, the previous work on keyphrase generation \citep{meng2017deep,chen2018keyphrase,Chen_Gao_Zhang_King_Lyu_2019,chan2019neural,yuan2020one} can be classified based on whether to generate the keyphrases individually (One2One, \citealt{meng2017deep}) or jointly (One2Seq, \citealt{yuan2020one}).
We propose to alleviate the severe negative effects of over-association in annotated gold queries. % caused by the one-to-many phenomenon in human dialogues.
To our knowledge, we are the first to investigate and alleviate such an issue for both tasks.
Besides, our findings can be beneficial to the keyphrase generation task as well.

In this work, we follow common practices \citep{komeili2022internet,WangSLMWTSY23} to conduct experiments on encoder-decoder structured models (T5 and BART).
Our preliminary study has shown that the over-association phenomenon can significantly affect the performance of both T5 and BART, which are pretrained with different corpora and training objectives. 
Therefore, we believe that models with alternative architectures, such as decoder-only based GPT2 \citep{radford2019language}, may also face similar challenges. 
Furthermore, given that both data-based weighting and model-based weighting strategies primarily concentrate on enhancing loss terms during fine-tuning without modifying the model structure, they can theoretically be adapted for use with various other types of models as well.

Currently, it is popular to adopt large language models (LLMs) to solve various natural language processing tasks using in-context learning. However, as shown in previous work \citep{WangSXS23,huang2023response}, even advanced models like InstructGPT \citep{NEURIPS2022_b1efde53} and ChatGPT\footnote{\url{https://openai.com/blog/chatgpt}} still demonstrate suboptimal performance compared to a much smaller task-specific model based on T5-base.
Nonetheless, given the exceptional capabilities of LLMs on text comprehension and generation \citep{achiam2023gpt}, we believe equipping LLMs with a task-specific query producer can further benefit the conversational query production and downstream response generation tasks \citep{xu2023small,shen2024hugginggpt}.

\section{Implications}
We investigate the under-explored over-association phenomenon and propose two instance-wise weighting strategies from different perspectives in this work.
Our findings not only help to build stronger query producers but also bring inspiration to future work in this field.

\emph{First}, it is important to carefully design the annotation guidelines for labeling search queries. Our preliminary study shows that the over-association phenomenon is common across current benchmarks, Wizard-of-Internet and DuSinc. 
Their guidelines can pay more attention to annotating responses but ignore the checking of labeled queries. 
As annotators may unconsciously perform reasoning with their background knowledge, a good response can sometimes come up with an unrelated query, e.g., ``\emph{the effect of eating orange}'' in Figure \ref{fig:example}, that a model struggles to predict.
Based on our conclusions, researchers can ask annotators to recognize the dialogue topic first before annotating and avoid introducing new concepts that are not necessary.

\emph{Second}, our proposed instance-wise weighting strategies can be used to improve the robustness of a model on noisy training data. 
As a dialogue system may serve users with different backgrounds, the topics of conversations can also vary across various domains. To ease the scarcity of annotations, a commonly used technique is data augmentation.
For instance, \citet{huang2023response} leverage dialogue responses to collect pseudo queries for semi-supervised training.
Similar to handling instances with high over-association degrees, our methods can help ease the negative impact of noisy instances as well by assigning smaller weights to these instances.

\section{Conclusion}
In this work, we have studied the over-association phenomenon in conversational query production, which is inherited from the open-ended nature of human dialogues.
We first systematically analyze the negative influence of over-association on standard text-generation models. Then, we designed several data-based and model-based weighting strategies that help to better train a model on the corpus suffering from over-association.
Experiments on two major datasets showed that all strategies and their combination indeed greatly alleviated the side effects. Our methods also effectively help solve data hunger and unfaithful issues raised by over-association.

\section*{Acknowledgments}
The project was supported by 
National Natural Science Foundation of China (No. 62276219),
Fujian Provincial Natural Science Foundation Project General Project (University Joint, No. 2024J01131070),  
and
the Public Technology Service Platform Project of Xiamen (No.3502Z20231043).
We thank reviewers for their insightful comments and also thank editors for their every effort.

\appendix
\section{Selection of $f$}
\label{sec:dev}

\begin{figure}[t]
\centering
    \begin{tikzpicture}[scale=0.8]
    \begin{axis}[
        legend style={
        at={(0.98,0.3)},
        cells={anchor=west}},
        xlabel=step,
        ylabel=Sum.,
        ]
    \addplot[sharp plot,color=red, mark=*] coordinates {
        (0, 131.65)(50, 136.19)(100, 139.26)(200, 138.83)(300, 140.03)(500, 140.01)
        };
    \addlegendentry{Uni. F$_1$}
    \addplot[dashed,color=green, mark=square*] coordinates {
        (0, 131.65)(50, 133.47)(100, 136.65)(200, 138.09)(300, 138.97)(500, 138.31)
        };
    \addlegendentry{BLEU-2}
    \addplot[densely dashed,color=blue, mark=triangle*] coordinates {
        (0, 131.65)(50, 139.21)(100, 139.23)(200, 140.24)(300, 140.18)(500, 140.29)
        };
    \addlegendentry{SBERT}
    \end{axis}
    \end{tikzpicture}
    \caption{Model performance at different training steps with different scoring functions on development set.}
    \label{fig:dev}
\end{figure}
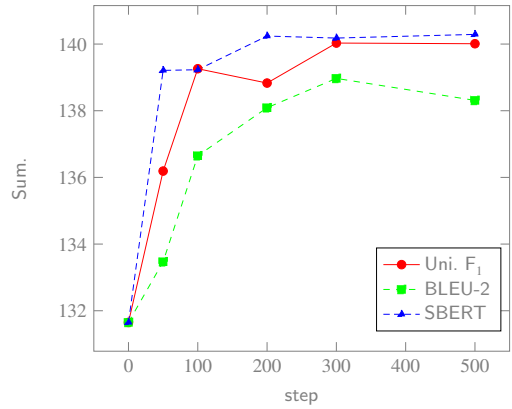

For scoring function $f$, we consider \emph{Uni. F$_1$}, \emph{BLEU-2} and semantic similarity calculated by a trained Sentence-BERT(SBERT)\footnote{We adopt mpnet-base model from \url{https://www.sbert.net/}}. Results are shown in Figure \ref{fig:dev}.
Using any scoring function significantly improves the model performance. We notice that \emph{Uni. F$_1$} and \emph{SBERT} perform competitively after 300 steps, but using \emph{SBERT} converges faster, which shows some benefits of considering deep semantic information. However, \emph{SBERT} takes much more time because of text embedding construction. 
Using \emph{BLEU-2} performs worst. It may be because \emph{BLEU-2} only focuses on precision while ignoring recall. 
% On the other hand, using \emph{BLEU-2} as $f$ performs worst on all metrics including \emph{BLEU-2} itself (40.52/39.75/39.38 when $f$ is \emph{Uni. F$_1$}/\emph{BLEU-2}/\emph{SBERT}). It may because \emph{BLEU-2} only focuses on precision ignoring recall. 
% This also shows that our generation-oriented approach tackles the root cause of model training instead of just improving the corresponding score participating in evaluation. 
In short, we select \emph{Uni. F$_1$} as $f$ considering performance and cost.

\section{Supplementary Evaluation Using ROUGE Scores}
\label{sec:rouge}

\begin{table*}[t] \small
\centering
\caption{Supplementary evaluation using \emph{ROUGE-1/2/L} on WoI.}
\begin{tabular}{lccc|ccc}
\toprule
\multirow{2}{*}{Model} &\multicolumn{3}{c|}{Wizard-of-Internet} & \multicolumn{3}{c}{DuSinc} \\
& ROUGE-1 & ROUGE-2 & ROUGE-L & ROUGE-1 & ROUGE-2 & ROUGE-L \\
\midrule
T5-base	& 47.38	& 27.00	& 46.25 & 68.18 & 57.03 & 66.30 \\
Data-weighting	& 49.19	& 28.85	& 48.19 & 68.84 & 57.92 & 66.96 \\
Model-wholeseq	& 51.13	& 28.42	& 49.58 & 70.23 & 59.95 & 68.43 \\
\makecell[l]{Data-weighting$\rightarrow$\\Model-wholeseq}	& 52.06	& 29.68	& 50.85 & 70.17 & 60.16 & 68.41 \\
\bottomrule
\end{tabular}
\label{tab:rouge}
\end{table*}

Table \ref{tab:rouge} gives the results of \emph{ROUGE-1/2/L} on WoI, which shows quite similar trends as the main test results.

\section{Annotation Guidelines}
\label{sec:guideline}
The human study aims to evaluate the quality of generated queries.
The evaluation is based on a 3-point scheme: 2 means flawless; 1 means having major flaw but with values; 0 means being completely wrong.

A soundness query should be factual and closely related to the current topic (especially the topic of the last few turn).
Take the following conversation as an example:

``\emph{User}: Have you ever seen the TV show Gilmore Girls? I just finished watching it for the third time!''

``\emph{Bot}: I have not ever seen it! But, one of my favorite actors is in it.''

``\emph{User}: Who is that? Alexis Bledel? She's one of my favorites.''

The gold query of the last turn is ``\emph{Jared Pad} Gilmore Girls'', which is a typical over-association query because the actor ``\emph{Jared Pad}'' is hard to be predicted from the dialogue context.

For the model predictions to this case, the following answers should be scored as 2:
\begin{itemize}
\setlength\itemsep{0em}
    \item \emph{Alexis Bledel Gilmore Girls}. It is highly related to the current topic about actors in ``Gilmore Girls''.
    \item \emph{actors in Gilmore Girls}. It helps the model reply to the question of user ``Who is that (actor)?''.
\end{itemize}

The following answers should be scored as 1:
\begin{itemize}
\setlength\itemsep{0em}
    \item \emph{Gilmore Girls}. It is not well related because the user tends to talk about favourite actor instead of the show at the last turn.
    \item \emph{Alexis Bledel}. It is not accurate enough that the retrieved documents may not contain knowledge about the role played by ``Alexis Bledel'' in the show.
\end{itemize}

The following answers should be scored as 0:
\begin{itemize}
\setlength\itemsep{0em}
    \item \emph{my favourite actor}. It does not contain meaningful information and can not help retrieving related knowledge to the conversation.
    \item \emph{Lionel Messi Gilmore Girls}. It is factually incorrect.
\end{itemize}


\bibliographystyle{cas-model2-names}

\bibliography{cas-refs}


% Author biography without author photo.
% Author biography. Author biography. Author biography.
% Author biography. Author biography. Author biography.
% Author biography. Author biography. Author biography.
% Author biography. Author biography. Author biography.
% Author biography. Author biography. Author biography.
% Author biography. Author biography. Author biography.
% Author biography. Author biography. Author biography.
% Author biography. Author biography. Author biography.
% Author biography. Author biography. Author biography.

% \bio{figs/pic1}
% Author biography with author photo.
% Author biography. Author biography. Author biography.
% Author biography. Author biography. Author biography.
% Author biography. Author biography. Author biography.
% Author biography. Author biography. Author biography.
% Author biography. Author biography. Author biography.
% Author biography. Author biography. Author biography.
% Author biography. Author biography. Author biography.
% Author biography. Author biography. Author biography.
% Author biography. Author biography. Author biography.

% \bio{figs/pic1}
% Author biography with author photo.
% Author biography. Author biography. Author biography.
% Author biography. Author biography. Author biography.
% Author biography. Author biography. Author biography.
% Author biography. Author biography. Author biography.

\end{document}